\documentclass[10pt,twocolumn,letterpaper]{article}

\usepackage{cvpr}
\usepackage{times}
\usepackage{epsfig}
\usepackage{graphicx}
\usepackage{mathtools}
\usepackage{amsmath}
\usepackage{amssymb}
\usepackage{bm}
\usepackage{algorithm}
\usepackage{algpseudocode}
\usepackage{tabularx}
\usepackage{subcaption}
\usepackage[skip=4pt,font=small]{caption}
\newcolumntype{Y}{>{\centering\arraybackslash}X}
\newcolumntype{s}{>{\hsize=.4\hsize}X}

\DeclareMathOperator\R{\mathbb{R}}
\DeclareMathOperator\M{\mathcal{M}}
\DeclareMathOperator\T{\mathcal{T}}
\DeclareMathOperator\sign{\text{sign}}
\DeclarePairedDelimiter{\norm}{\lVert}{\rVert}

\DeclarePairedDelimiter{\floor}{\lfloor}{\rfloor}

\usepackage[pagebackref=true,breaklinks=true,letterpaper=true,colorlinks,bookmarks=false]{hyperref}

 \cvprfinalcopy 

\def\cvprPaperID{2844} 

\ifcvprfinal\fi
\begin{document}

\title{Geometric robustness of deep networks: analysis and improvement}

\author{Can Kanbak, Seyed-Mohsen Moosavi-Dezfooli, Pascal Frossard\\
\'Ecole Polytechnique F\'ed\'erale de Lausanne\\
{\tt\small \{can.kanbak, seyed.moosavi, pascal.frossard\} @epfl.ch}
}

\maketitle

\begin{abstract}
   Deep convolutional neural networks have been shown to be vulnerable to arbitrary geometric transformations. However, there is no systematic method to measure the invariance properties of deep networks to such transformations. We propose ManiFool as a simple yet scalable algorithm to measure the invariance of deep networks. In particular, our algorithm measures the robustness of deep networks to geometric transformations in a worst-case regime as they can be problematic for sensitive applications. Our extensive experimental results show that ManiFool can be used to measure the invariance of fairly complex networks on high dimensional datasets and these values can be used for analyzing the reasons for it. Furthermore, we build on Manifool to propose a new adversarial training scheme and we show its effectiveness on improving the invariance properties of deep neural networks.\footnote{To encourage reproducible research, the code of our method will be later published.} 
\end{abstract}

\section{Introduction}
Although convolutional neural networks (CNNs) have been largely successful in various applications, they have been shown to be quite vulnerable to additive adversarial perturbations~\cite{szegedy_intriguing_2014, fgs, deepfool} which can negatively affect their applicability in sensitive applications such as autonomous driving ~\cite{evtimov_robust_2017}.  Deep networks have also been shown to be vulnerable to rigid geometric transformations \cite{manitest, goodfellow_measuring_2009}, which are more natural than additive perturbations: they can simply represent the change of the viewpoint of an image. Therefore, invariance to such transformations is certainly a key feature in practical vision systems. In this paper, we focus on studying the robustness of deep networks to geometric transformations in the worst-case regime as these can be quite problematic for sensitive applications. We approach this problem by searching for minimal 'fooling' transformations, \ie , transformations that change the decision of image classifiers, and we use these transformed examples to measure the invariance of a deep network. We further show that fine-tuning on such worst-case transformed examples can improve the invariance properties of deep image classifiers. Our main contributions are as follows:
\begin{figure}[t]
  \center
  \includegraphics[width=0.3\textwidth]{{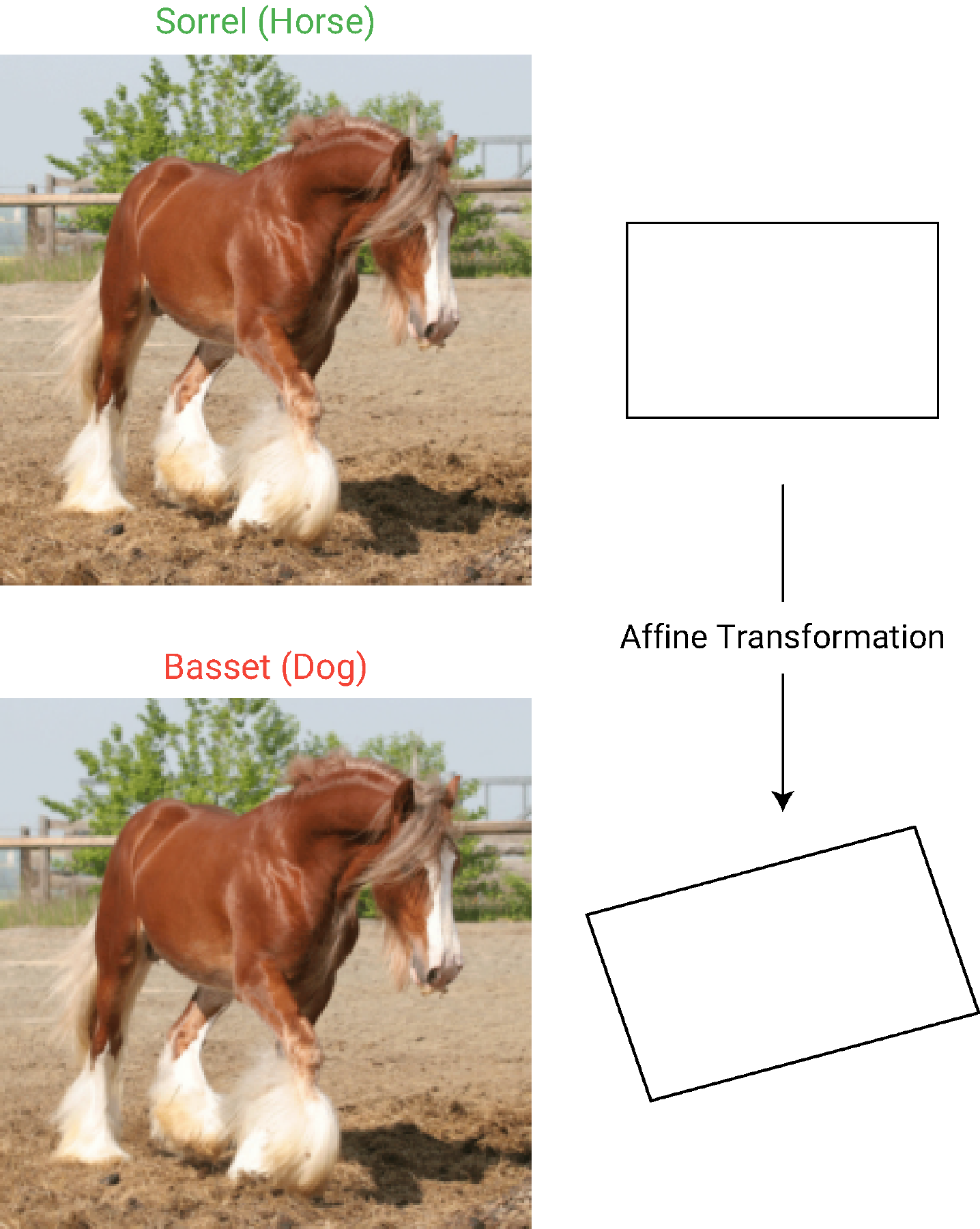}}
  \caption{An example of a worst-case 'fooling' affine transform for AlexNet \cite{krizhevsky_imagenet_2012}. While the image on top is correctly classified as sorrel (a type of horse), a small transformation in the bottom image can cause it to be classified as basset (a type of dog), even though the change in the image is imperceptible. }
  \label{}
\end{figure}
\begin{itemize}
\item We propose a scalable algorithm, ManiFool, for finding small worst-case transformations and define a measure to compare the invariance properties of different networks.
\item As far as we know, we perform the first quantitative study on the robustness of deep networks to geometric transformations that are trained on a large scale dataset, \ie , ImageNet, and show that these networks are susceptible to small and sometimes imperceptible \textit{transformations}.
\item We use the ManiFool algorithm to perform adversarial training using geometric transformations and show that it actually improves the invariance of deep networks.
\end{itemize}
The adversarial examples are first introduced in \cite{szegedy_intriguing_2014}. Since then, many methods to find additive adversarial perturbations have been proposed such as \cite{deepfool, fgs, carlini_evaluating_2017}. Other types of adversarial examples are later found in \cite{baluja_adversarial_2017, sabour_adversarial_2016}. The work \cite{szegedy_intriguing_2014} also introduces the concept of adversarial training to increase the accuracy of networks. The authors in \cite{fgs} later show that adversarial training can also be used for increasing the robustness of networks against adversarial examples constructed by additive perturbations. 

The vulnerability of CNNs against geometric transformations, on the other hand, has been studied in~\cite{lenc_understanding_2015} and \cite{soatto_visual_2016} that analyze image and visual representations to find theoretical foundations of transformation invariant features. The work in \cite{bakry_digging_2016} uses the information about human visual system to understand and improve the transformation invariance. In addition, several practical solutions have been suggested for improving the invariance characteristics. One approach is to modify the layers of the networks; \eg, pooling layer\cite{dai_deformable_2017} or convolutional layers \cite{shen_transform-invariant_2016}. Another method is to add modules to the network, like the spatial transformer networks \cite{jaderberg_spatial_2015}. Even though these works focus on improving the invariance, they do not offer methods for measuring invariance properties of classification architectures. This problem is the main focus of \cite{goodfellow_measuring_2009}, where the invariance is measured by using the firing rates of neurons in the network for one dimensional transformations. On the other hand, the authors of \cite{fawzi_measuring_2016} propose a probabilistic framework for estimating the robustness of a classifier by using a Metropolis algorithm to sample the set of transformations. Lastly, another approach is given by Manitest \cite{manitest}, where the invariance is measured using the geodesic distances on the manifold of transformed images. In this work, we also use a manifold-based definition of invariance and propose a new scalable algorithm for evaluating invariance in more complex networks and improving it by fine-tuning.
\section{Preliminaries}


In this section, we briefly introduce the mathematical tools that we will use to measure the robustness of deep networks to geometric transformations.

Let $\T$ be a Lie group of geometric transformations such as rotations or projective transformations. Then, let $\tau\in\T$ be a function $\tau:\R^2 \rightarrow \R^2$ in this group. For 2D images, $\tau$ can be seen as a bijection that maps the points of an image to the points of another image. More precisely, let an image $I$ be defined as a square integrable function $I:\R^2 \to \R$. The action of $\tau$ on $I$ can be represented as $I_\tau(x,y)$. This can also be seen as a function that maps the Lie group $\T$ to the image space, which we denote by $\psi^{(I)}(\tau): \T \to L^2$, where $L^2$ is the space of square integrable functions. 
A transformation $\tau$ can be represented by as many parameters as the dimensionality of $\T$. For example, the rotation angle can be used to parameterize the rotation group. These parameters can be grouped in a vector $\bm{\theta}$, where each element represents one of the parameters of $\tau$.

For a given Lie group $\T$, we need to define a metric $d(\tau_1, \tau_2): \T \times \T \to \R$ to measure the actual effect of transformations on images. A naive metric would consist in measuring the $\ell^2$ distance between the parameter vectors of $\tau_1$ and $\tau_2$. This however is not a useful metric since it does not take into account the different nature of the transformation parameters such as rotation angle or scale. The metric should rather depend on the image as well as the transformations. However, another metric such as the squared $L^2$ distance $d_I(\tau_1, \tau_2) = \norm{I_{\tau_1} - I_{\tau_2}}^2_{L^2}$ still does not fully capture the properties of the transformation even if it depends on the image(see \cite{manitest} for an illustrative example). Thus, a better metric should be able to capture the intrinsic geometric structure of the transformed images.

 One such metric is to the length of the shortest curve between $\tau_1,\tau_2 \in \T$, \ie , the \emph{geodesic distance}. This metric, however, requires a Riemannian metric to be defined for $\T$. In this case, the Riemannian metric can be acquired by mapping $\T$ to the set of transformed versions of image $I$, \ie , $\M(I)=\left\{I_\tau : \tau \in \T\right\}$. This set forms a differentiable manifold called the image appearance manifold (IAM) following the works of  \cite{wakin_multiscale_2005,kokiopoulou_minimum_2009} and it inherits a Riemannian metric from its ambient space, $L^2$. From \cite{manitest}, it can be seen that the Riemannian metric on $\T$ can be chosen accordingly, such that the length of a curve on $\T$, $\gamma(t):[0,1] \to \T$, will be equal to the length of the mapped curve on $\M(I)$, $I_{\gamma(t)}:[0,1] \to \M(I)$. Thus, the geodesic distance between $\tau_1,\tau_2 \in \T$ is equal to the geodesic distance between $I_{\tau_1},I_{\tau_2} \in \M(I)$. Then, for $\tau_1,\tau_2 \in \T$, the transformation metric can be finally defined as
\begin{equation}\label{d_i}
	\begin{split}
	d_I(\tau_1,\tau_2) = &\min_{\gamma:[0,1] \to \M(I)} L(\gamma) \\
	&\text{s.t.} \, \gamma(0) = I_{\tau_1},\gamma(1) = I_{\tau_2},
	\end{split}
\end{equation}
where $L(\gamma)$ is the length of the curve $\gamma$. This metric both depends on the transformed image and also takes into account the geometric properties of the transformation set $\T$. In order to measure the action of a transformation $\tau$, we therefore use the distance in Eq. \eqref{d_i} --between the transformation and the identity transformation $e$, \ie , $d_I(e,\tau)$. To compare the effect of transformations on different images, we normalize $d_I(e,\tau)$ by the norm the image, that is
\begin{equation} \label{metric}
\tilde{d}_I(e,\tau) = \frac{d_I(e,\tau)}{\norm{I}_{L^2}}.
\end{equation}

\section{Robustness to geometric transformations}\label{proform}

As we have now chosen our metric as \eqref{metric}, our approach to measure robustness of classifiers can be formalized as follows. Let $k$ be the given classifier, $I$ an image and  $\T$ the set of transformations we are interested in. As with \cite{manitest}, we define the invariance metric as the minimal normalized distance between the identity transformation and a transformation that leads to misclassification. Hence, the invariance measure is denoted as
\begin{equation}
\Delta_{\T}(I,k) = \min_{\tau \in \T}\tilde{d}(e,\tau) \quad \text{subject to} \quad k(I) \not= k(I_\tau),
\end{equation}
Note that this definition is similar to those used in works on adversarial perturbations such as \cite{deepfool}\cite{goodfellow_measuring_2009}. However, in our case, instead of looking at minimal fooling additive perturbations, we are interested in the minimal or worst-case geometric transformation. Thus, for a probability distribution $\mu$ on the set of images, the global invariance score of the classifier $k$ to transformations in $\T$ is defined as
\begin{equation}
\rho_{\T}(k) = \mathbb{E}_{I \sim \mu}\Delta_{\T}(I,k).
\end{equation}
 As the underlying probability distribution of the images is generally unknown, the invariance score of the classifier is calculated using the empirical average of the estimated invariance over a set of images:
\begin{equation}\label{rhoest}
\hat{\rho}_{\T}(k) = \frac{1}{m}\sum_{j=1}^m\tilde{d}_{I_j}(e,\hat{\tau}).
\end{equation}

One can also define the invariance of a classifier to a random transformation by again using the metric $\tilde{d}(e,\tau)$. In this case, we use the probability of misclassification of transformed images under random transformations with given geodesic scores to measure the invariance. This can be defined it as
\begin{equation}\label{rest}
r_{\T}(k) = \min r \; \text{s.t.} \; \underset{I,\tau}{\mathbb{P}}(k(I_\tau)\not=k(I) \mid  d_I(e,\tau)=r ) \ge 0.5.
\end{equation}
Note that $0.5$ is chosen as a threshold here, but other thresholds can be used for defining the invariance to random transformations.

In practice, $\hat{\rho}_{\T}(k)$ in \eqref{rhoest} is computed using the algorithm described in Section 4 to find a small transformation $\hat{\tau}$ that can fool the image and using the geodesic distance of this transformation, $\tilde{d}_I(e,\hat{\tau})$, to estimate  $\Delta_{\T}(I,k)$. The expectation can then be calculated using multiple images sampled from a dataset. On the other hand, the estimation of $r_{\T}$, $\hat{r}_{\T}(k)$, is computed by sampling the set $\M_r  = \left\{I_\tau: \tau \in \T,  \: \tilde{d}_I(e,\tau) = r\right\}$  for increasing $r$ and computing the misclassification percentage of the corresponding transformed samples.

\section{ManiFool}
\begin{figure}
  \small
  \centering
    \includegraphics[width=0.4\textwidth , keepaspectratio]{{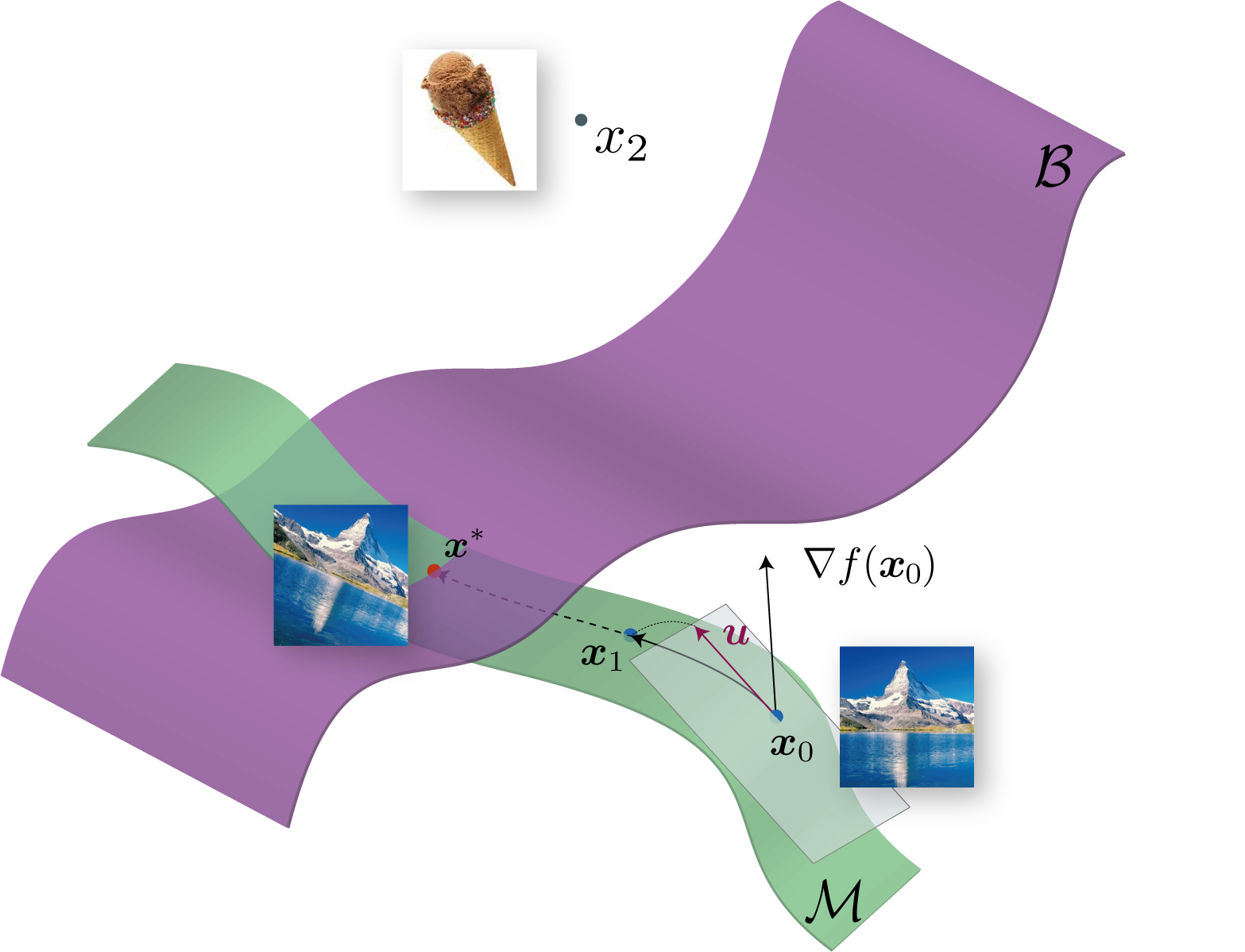}}
    \caption{Illustration of Algorithm \ref{alg:manifool}. Assume $\M$ is the manifold of transformed images for the input $\bm{x}_0$ and $\mathcal{B}$ is the decision boundary of classifier $f$. The algorithm iteratively moves towards the decision boundary. The first iteration is shown where the movement direction $\bm{u}$ is found by projecting $\nabla f$ to the tangential space of $\M$ and the next image $\bm{x}_1$ is found by mapping $u$ back onto the manifold }
    \label{fig:manifool}
\end{figure}

In this section, we first introduce the ManiFool algorithm to find a small fooling transformation $\hat{\tau}$ for binary and multiclass classifiers. We then present a method to measure the geodesic distance $\tilde{d}_{I_j}(e,\hat{\tau})$ to compute the invariance score in \eqref{rhoest}. The main idea of the ManiFool algorithm is simply to iteratively move from an image sample towards the decision boundary of the classifier where the classification decision changes, while staying on the transformation manifold. Each iteration is then composed of two steps:  choosing the movement direction and mapping this movement onto the manifold. The iterations will continue until the algorithm reaches the decision boundary and finds a fooling transformation example. An illustration of the algorithm can be found in Figure \ref{fig:manifool} and a more detailed description of the algorithm is given in the following section.

\subsection{ManiFool for binary classifiers}
Since a multiclass classifier can be thought as a combination of multiple binary classifiers, we first start from the binary classifier case. We consider from now on discrete versions of images, denoted as $\bm{x} \in \R^n$ where $n$ is the number of pixels of $\bm{x}$. The binary classifier is defined as $k(x) = \sign(f(x))$, where $f : R^n \to \R$ is an arbitrary differentiable classification function. Without loss of generality, we can assume that the original label of the image is 1.

Let $\bm{x}^{(i)}$ be the image at the start of iteration $i$ and let iterations start with $i=0$. The first step for any iteration consists in finding the movement direction. Since we assumed that the original label is 1, $f(\bm{x}^{(0)}) > 0$ for the input image $\bm{x}^{(0)}$. Thus, to reach the decision boundary where $f(\bm{x}) = 0$ while following the shortest path, we need to choose the direction which maximally reduces $f(\bm{x})$, which is the opposite of the gradient of $f$,  $-\nabla f(\bm{x})$. However, since we want to stay on the set of transformed images, we restrict the classifier to the image appearance manifold, $\M(\bm{x})$, as $f_{\mid \M}: \M(\bm{x}) \to \R$, and use its gradient $\nabla f_{\mid \M}$. At point $\bm{x}^{(i)}$, this gradient can be acquired simply by projecting $\nabla f(\bm{x}^{(i)})$ onto the tangent space of $\bm{x}^{(i)}$, $T_{\bm{x}^{(i)}}\M$ \cite{absil_optimization_2008}. This projection is done using the pseudoinverse operator as
\begin{equation}
\bm{u} = -\bm{J}_{\bm{x}^{(i)}}^+\nabla f(\bm{x}^{(i)}) = -(\bm{J}_{\bm{x}^{(i)}}^T\bm{J}_{\bm{x}^{(i)}})^{-1}\bm{J}_{\bm{x}^{(i)}}^T\nabla f(\bm{x}^{(i)}).
\end{equation}
where $\bm{J}_{\bm{x}^{(i)}}$ is the Jacobian matrix, whose columns form the basis of the tangential space and  $\bm{u} \in T_{\bm{x}^{(i)}}\M$ is the projection of  $\nabla f(\bm{x}^{(i)})$, \ie , our movement direction for this iteration.


After finding the movement direction, the next step is to map $\bm{u}$ onto $\M(\bm{x})$. This step depends heavily on the transformation set. As we want to minimize the geodesic distance, the natural choice of mapping would be to use the exponential map for $\M(\bm{x})$ since it follows the geodesics \cite{tu_differential_2017}. If an exponential map is readily available for $\M$  and does not have high computational complexity, it can be used. However, for most transformation sets, this does not hold and a retraction is used instead. Here, we will talk about one such retraction for the set of projective transformations and its subsets. Different retractions can be defined for other Lie groups.

\begin{algorithm}[t]
\caption{ManiFool for binary classifiers}
\label{alg:manifool}
\begin{algorithmic}[1]
\Require Image $\bm{x}$, classifier $f$
\Ensure Transformation $\hat{\tau}$

\State Initialize with $\bm{x}^{(0)} \leftarrow \bm{x}$, $i \leftarrow 0$.
\While{$f(\bm{x}^{(i)})>0$ }
	\State $\hat{\bm{u}} \gets -\bm{J}_{\bm{x}^{(i)}}^+\nabla f(\bm{x}^{(i)})$
	\State $\bm{u}^{(i)} \gets \lambda_i \frac{\hat{\bm{u}}}{\norm{\hat{\bm{u}}_{l_n}}} + \gamma\bm{u}^{(i-1)}$
	\State $\tau_i \gets \exp\left(\sum_j u_jG_j\right)$
	\State $\bm{x}^{(i+1)} \gets \bm{x}^{(i)}_{\tau_i}$
	\State $i \gets i+1$
\EndWhile \\
\Return  $\hat{\tau} = \tau_0 \circ \tau_1 \circ \dots \tau_i$
\end{algorithmic}
\end{algorithm}

The retraction in our implementation uses the matrix representation of projective transformations, where the matrix exponential forms a map from $T_e\T$ to $\T$. Let $\bm{y} \in \M(\bm{x})$, $\bm{u} \in T_{\bm{y}}\M(\bm{x})$ and $G_i$ be the basis of $T_e\T$, which are also called the generators of $\T$. The retraction we use can be summarized as mapping $\bm{u}$ to $T_e\T$ by using the generators, then mapping it to the matrix Lie group $\T$ and lastly, mapping it back to $\M(x)$ by using $\psi^{\bm{x}^{(i)}}$. More formally, the retraction at the point $\bm{y}$, $R_{\bm{y}}: T_{\bm{y}}\M \to \M$ can be written as
\begin{equation}\label{retr}
R_{\bm{y}}(\bm{u}) = \psi^{(\bm{y})} \left(\exp\left(\sum_j u_jG_j\right)\right).
\end{equation}
The image for the next iteration of the algorithm is thus written as
\begin{equation}
\bm{x}^{(i+1)} = R_{\bm{x}^{(i)}}(\bm{u}) = \bm{x}^{(i)}_{\tau_i},
\end{equation}
where $\tau_i$ is the transformation represented by
\begin{equation}
\tau_i = \exp\left(\sum_j u_jG_j\right).
\end{equation}

Lastly, the label of the generated image is checked. If $k(\bm{x}^{(i+1)}) = 1$, the algorithm continues with the next iteration, this time starting from $\bm{x}^{(i+1)}$. Otherwise, the algorithm has finished successfully and the transformation that generated $\bm{x}^{(i+1)}$ is found as
\begin{equation}
\hat{\tau} = \tau_0 \circ \tau_1 \circ \dots \tau_i.
\end{equation}

The algorithm is summarized on Algorithm \ref{alg:manifool}. Overall, it should be noted that our algorithm is closely related to manifold optimization techniques, particularly to line-search methods. The  convergence analysis of such methods can be found for example in \cite{absil_optimization_2008}. Using this analogy, choosing the movement direction has changed by including line-search and momentum terms, since it has been seen empirically that they improve the accuracy and reduce the chance of converging to a local minimum. This new direction term can be written as
\begin{equation}
\bm{u}^{(i)} = -\lambda_i\frac{\bm{J}_{\bm{x}^{(i)}}^+\nabla f(\bm{x}^{(i)})}{\norm{\bm{J}_{\bm{x}^{(i)}}^+\nabla f(\bm{x}^{(i)})}} + \gamma\bm{u}^{(i-1)},
\end{equation}
where $\lambda_i$ is a step size term that is chosen to maximize the decrease in $f$ in each step and $\gamma$ is the constant momentum parameter.

\subsection{ManiFool for multiclass classifiers}
The most common scheme used in multiclass classifiers is one-vs-all, which serves as a basis for our method. In this scheme, the classifier function has $c$ outputs where $c$ is the number classes. Thus, the function is defined as $f : \R^n \to \R^c$ and the classification is performed as:
\begin{equation}
	k(\bm{x}) = \arg\max_k f_k(\bm{x}),
\end{equation}
where $f_k$ is the output of $f$ that corresponds to the $k^{\text{th}}$ class. Let $l_x = k(\bm{x}^{(0)})$, where $\bm{x}^{(0)}$ is the input image to the algorithm. Then we can define $c-1$ binary classifiers for $l \not= l_x$ as:
\begin{equation}
 g_l(\bm{x}) = f_{l_x}(\bm{x}) - f_l(\bm{x}).
\end{equation}
Since $l_x  = \arg\max_k f_k(\bm{x}^{(0)})$,  $f_{l_x}(\bm{x^{(0)}}) > f_l(\bm{x^{(0)}})$ for all $l \not= l_x$ and thus $g_l(\bm{x}^{(0)}) > 0$.

As we now have $c-1$ binary classifiers, we can get $c-1$ examples of fooling transformations by using each binary classifier as input to the binary ManiFool from Algorithm \ref{alg:manifool}. When the algorithm is used with $g_l$, it stops iterating when $k(\bm{x}^{(i)}) \not= l_x$, instead of stopping when $g_l(\bm{x}^{(i)}) < 0$, because the classifier we are trying to fool is $k$, and not $g_l$.  Let $\tau_l$ be the output transformation  when $g_l$ is used as the input to the binary ManiFool. As we now have a fooling transformation example for each class, the class with the smallest transformation by using  the geodesic score from \eqref{metric} can be chosen as
\begin{equation}
l_{\min} = \arg\min_{l \not= l_x} \tilde{d}_{\bm{x}^{(0)}}(e,\tau_l),
\end{equation}
and the algorithm will output the corresponding transformation, $\tau_{l_{\min}}$.

\begin{algorithm}[t]
\caption{ManiFool for multiclass classifiers}
\label{alg:manifool_multi}
\begin{algorithmic}[1]
\Require Image $\bm{x}$, classifier $f$
\Ensure Transformation $\hat{\tau}$

\State Initialize with $\bm{x}^{(0)} \leftarrow \bm{x}$, $l_x \gets k(\bm{x}^{(0)})$.
\For{$l \not= l_x$}
	\State $ g_l \gets f_{l_x} - f_l$
	\State $i \leftarrow 0$
	\While{$k(\bm{x}^{(i)}) = l_x$}
		\State $\hat{\bm{u}} \gets -\bm{J}_{\bm{x}^{(i)}}^+\nabla g_l(\bm{x}^{(i)})$
		\State $\bm{u}^{(i)} \gets \lambda_i \frac{\hat{\bm{u}}}{\norm{\hat{\bm{u}}_{l_n}}} + \gamma\bm{u}^{(i-1)}$
		\State $\tau_i \gets \exp\left(\sum_j u_jG_j\right)$
		\State $\bm{x}^{(i+1)} \gets \bm{x}^{(i)}_{\tau_i}$
		\State $i \gets i+1$
	\EndWhile
	\State $\hat{\tau}_l \gets \tau_0 \circ \tau_1 \circ \dots \tau_i$
\EndFor
\State $l_{min} \gets \arg\min_{l \not= l_x} \tilde{d}(e,\tau_l)$ \\
\Return  $\hat{\tau} = \tau_{l_{min}}$
\end{algorithmic}
\end{algorithm}

The complexity of the algorithm depends on multiple factors, including the properties of the input classifier, input image and parameters such as $\gamma$. For example, as a separate single target ManiFool is run for each output class, the complexity depends heavily on $c$, the number of classes. Thus, to reduce complexity, only the most probable $\hat{c}$ classes are used in the algorithm, which are the $\hat{c}$ classes with highest $f_l(\bm{x}_0)$ excluding $l_x$. The complexity is also highly dependent on the number of iterations $N_{\text{it}}$ for each single target ManiFool. Although these cannot be known exactly before running the algorithm, they are bounded by the maximum number of iterations $N_{\max}$. Thus, the complexity increases linearly with $N_{\max}$, which should be chosen carefully in order not to unnecessarily increase the complexity.

\subsection{Measuring robustness to geometric transformations}\label{trans_measure}

\begin{figure}
  \small
  \centering
    \includegraphics[width=0.4\textwidth , keepaspectratio]{{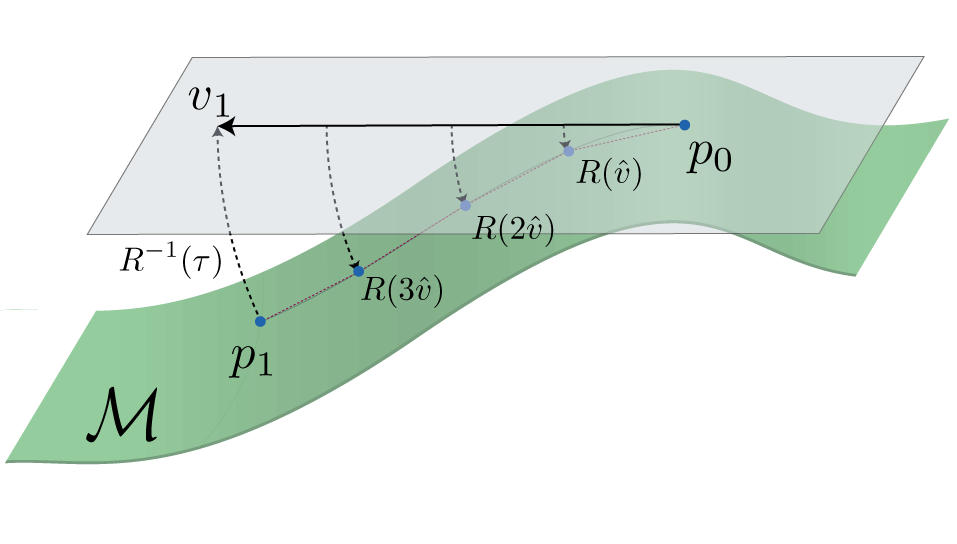}}
    \caption{ Illustration of the distance measurement. Assume $\M$ be a manifold and $p_0,p_1 \in \M$. We estimate the geodesic distance between these points using the direct path, by first mapping $p_1$ to the tangential space of $p_0$ as $v_1$ vector, dividing $v_1$ into smaller vectors and remapping these back onto the manifold. Then, $\hat{d}(p_0,p_1)$ from \eqref{dd_metric} is given by the sum of distances between these points, as the length of the grey curve.}
    \label{fig:dd}
\end{figure}

Although we have utilized geodesic distance $d_{\bm{x}}(e,\tau)$ in the above methods, we have not shown how it can be computed. One possible method is to use Fast Marching Method (FMM), which progressively calculates the distance of points on a grid on the manifold from a reference point  \cite{sethian_fast_1996}. However, the complexity of this algorithm increases exponentially with the manifold dimension, and may rapidly become too complex. Thus, we propose a different method, by assuming that the geodesic path to the target node is direct, and we estimate the geodesic distance using this direct path. Let $p_0,p_1 \in \M$, and $v_1 = R_{p_0}^{-1}(p_1)$ where $R$ is a retraction. Then, the direct path from $p_0$ to $p_1$ can be defined as
\begin{equation}
\gamma(t) = R_{p_0}(tv_1), \quad  t \in [0,1].
\end{equation}

\begin{figure*}[t!]
\centering
 	\begin{subfigure}[t] {0.4\textwidth}
   		\includegraphics[width=\textwidth , keepaspectratio]{{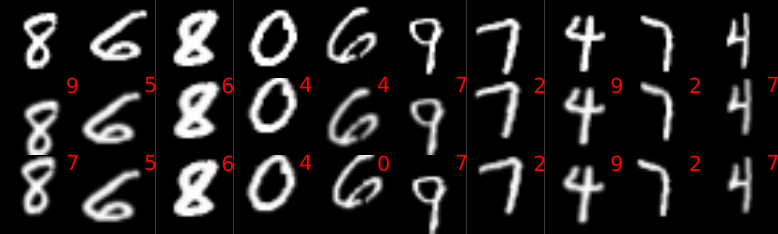}}
    		\caption{Translations}
    	\end{subfigure}
    	\hspace{5mm}
    	\begin{subfigure}[t] {0.4\textwidth}
   		\includegraphics[width=\textwidth , keepaspectratio]{{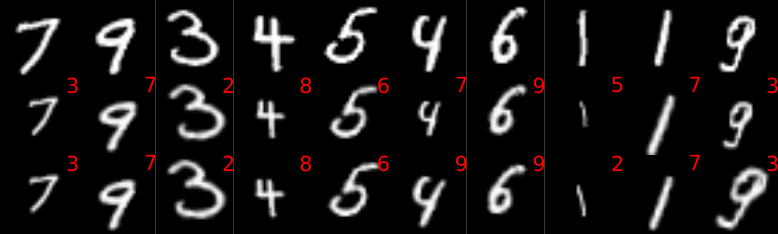}}
    		\caption{Similarity transformations}
    	\end{subfigure}
    	\caption{Examples of MNIST images transformed using outputs of ManiFool and Manitest \cite{manitest} for translation and similarity sets. Top rows show the original images, the middle rows show the outputs from ManiFool and the bottom row shows the output of Manitest. The red numbers indicate the new output labels of the transformed images. }
	\label{fig:mfvsmt}
\end{figure*}

For a chosen step-size $\eta$, $v_1$ can be divided into parts as $\hat{v} = \eta\frac{v_1}{\norm{v_1}}$. Then for $N = \floor{\frac{\norm{v_1}}{\eta}}$, the distance can be estimated as
\begin{equation}\label{dd_metric}
\hat{d}(p_0,p_1) = \sum_{i=1}^N\norm{R(i\hat{v})-R((i-1)\hat{v})}_{L^2} + \norm{p_1-R(N\hat{v})}_{L^2},
\end{equation}
which is the sum of $L^2$ distances on the sampled direct path. An illustration is given on Figure \ref{fig:dd}. In our case, we estimate $d_{\bm{x}}(e,\tau)$ as $\hat{d}(\bm{x},\bm{x}_\tau)$, since from \eqref{d_i}, the distance between the transformations on $\T$ is equal to the distance between their transformed counterparts.

\section{Experimental Results}

We now test our algorithm on convolutional neural network architectures. In these experiments, the invariance score for minimal transformations, defined in \eqref{rhoest}, is calculated by finding fooling transformation examples using ManiFool for a set of images, and computing the average of the geodesic distance of these examples. On the other hand, to calculate the invariance against random transformations, we generate a number of random transformations with a given geodesic distance $r$ for each image in a set and calculate the misclassification rate\footnote{As we consider the invariance of a network, we define misclassification as a change in label, \ie , if $k(\bm{x}) \not= k(\bm{x}_\tau)$ for the transformed image $\bm{x}_\tau$} of the network for the transformed images. A random transformation is created by sampling the unit sphere of $T_e\T$ and increasing the magnitude of this vector until the corresponding transformation has score equal to $r$, \ie , $d(e,R(\alpha v)) = r$ where $v \in T_e\T$ is the sampled vector with $\norm{v} = 1$ and $\alpha > 0$ is a scaling factor. A more detailed explanation of how the random transformation is sampled can be found in supp. material. The misclassification rate is calculated for different $r$ to see the performance of the network on different levels of perturbation and to get $\hat{r}_{\T}$ from \eqref{rest}.  In every case, the discrete images after transformation are obtained using bilinear interpolation; they further have the same size as the original image with zero-padding boundary conditions when necessary.

\subsection{Performance of ManiFool}

\begin{figure*}[ht]
  \small
  \centering
    \includegraphics[width=0.8\textwidth , keepaspectratio]{{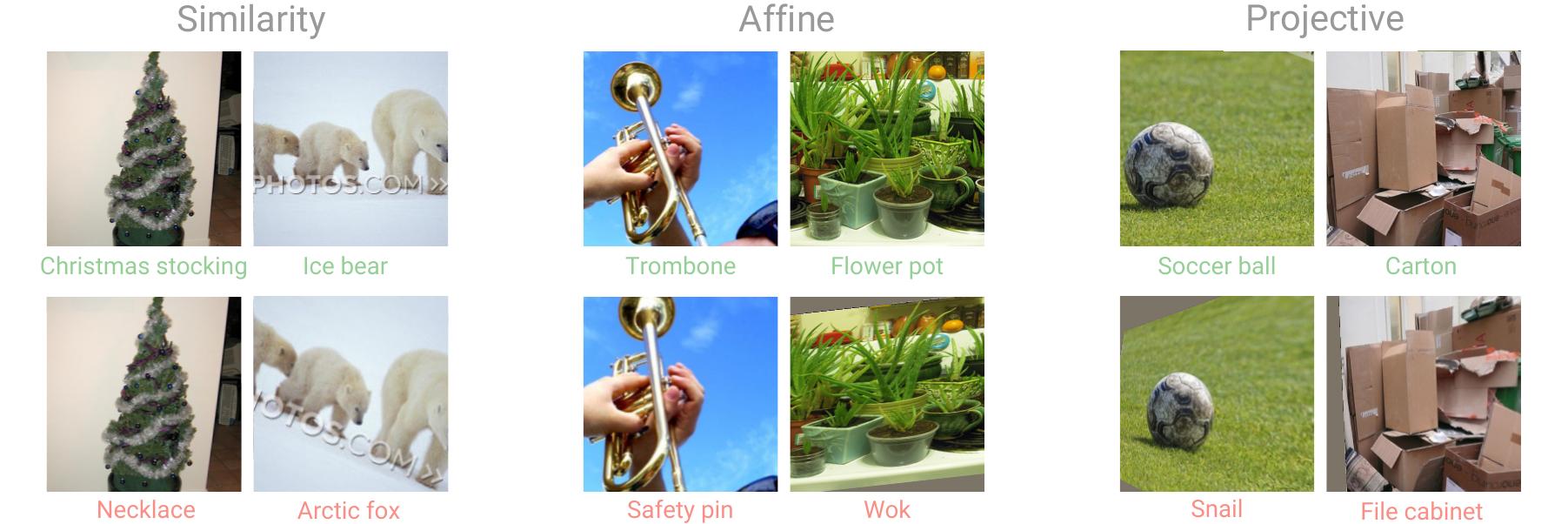}}
    \caption{Examples of ILSVRC2012 images transformed using outputs of ManiFool using ResNet18 for similarity, affine and projective sets.Top row shows the original image while the bottom row shows the transformed image. The texts bottom show the output labels of the images at top and bottom respectively. More examples are found in supp. material.}
    \label{fig:imagenet}
\end{figure*}

The first experiment compares the ManiFool algorithm with  Manitest \cite{manitest},  to evaluate the performance of the algorithm in terms of speed and accuracy. For this comparison, on top of calculating the invariance score using ManiFool with $N_{\max} = 50$, we also do the same thing with Manitest, \ie , we use it to find fooling transformation examples for the set of images and use these transformations to measure invariance. The comparison is done using 1000 images from the MNIST \cite{mnist} training dataset and a baseline CNN with two $5\times 5$ layers with 32 and 64 feature maps respectively with ReLU nonlinearity and $2\times 2$ max pooling. In both cases, the geodesic distance is calculated using \eqref{dd_metric} to be comparable. Some of the transformed images generated during the experiment can be seen in Figure \ref{fig:mfvsmt} and Table \ref{tab:mtvsmf} reports the invariance score $\hat{\rho}_{\T}$ and the running time for both methods.

Manitest uses the fast marching method to find the transformation that changes the label. This requires measuring the distance of all the transformed images on a grid over parameters and evaluate the classifier until it reaches a fooling transformation. Thus, it is guaranteed to find the minimal transformation in the discretized search space and it is more accurate than ManiFool in this regard. However, as it can be seen on Table \ref{tab:mtvsmf}, it is more complex than ManiFool, especially as the dimension of the manifold is increased. Thus, it is not scalable for high dimensional manifolds. On the other hand, while not being as accurate, ManiFool is less complex in these situations. For example, Table \ref{tab:mtvsmf} shows that the complexity of ManiFool does not closely depend on the dimensionality of $\M(\bm{x})$. Thus, ManiFool can be used for measuring the invariance of more complex networks such as the state-of-the-art architectures used with ImageNet database. Because of this, we have only used ManiFool to compare networks in the following experiments.

\begin{table}[t]
\centering \small
\renewcommand{\arraystretch}{1}
\begin{tabularx}{0.45\textwidth}{ |X|s|s|s|s| }
 \hline
 & \multicolumn{2}{c|}{ManiFool} & \multicolumn{2}{c|}{Manitest}\\ \hline
 Transformation & \raggedright $\hat{\rho}_{\T}$  & \raggedright time  & \raggedright  $\hat{\rho}_{\T}$  & \raggedright time  \tabularnewline
  \hline
   T ($d=2$) & 1.68 & 2.6 s & 1.54 & 2.7 s \\ \hline
   R+T ($d=3$)  & 1.40 & 3.6 s & 1.33 & 23.9 s \\ \hline
   S+T ($d=3$) & 1.41  & 6.2 s &  1.32 &  34.6 s  \\ \hline
   T+R+S ($d=4$) & 1.26  & 3.1 s &  1.25 & 29.5 s  \\ \hline
\end{tabularx}
\caption{Comparison of Manitest and ManiFool for different transformation sets on MNIST dataset. In the table, T, R and S stand for translation, rotation and scaling respectively; while $d$ represents the number of dimensions of the transformation groups. The time column lists the average time required to compute one sample. The experiment was done using a baseline CNN with 2 convolutional layers. These times are computed on a server with 2 Intel Xeon CPU E5-2680 v3 without GPU support.}
\label{tab:mtvsmf}
\end{table}

\begin{figure*}[ht]
\centering
 	\begin{subfigure}[t] {0.45\textwidth}
   		\includegraphics[width=\textwidth , keepaspectratio]{{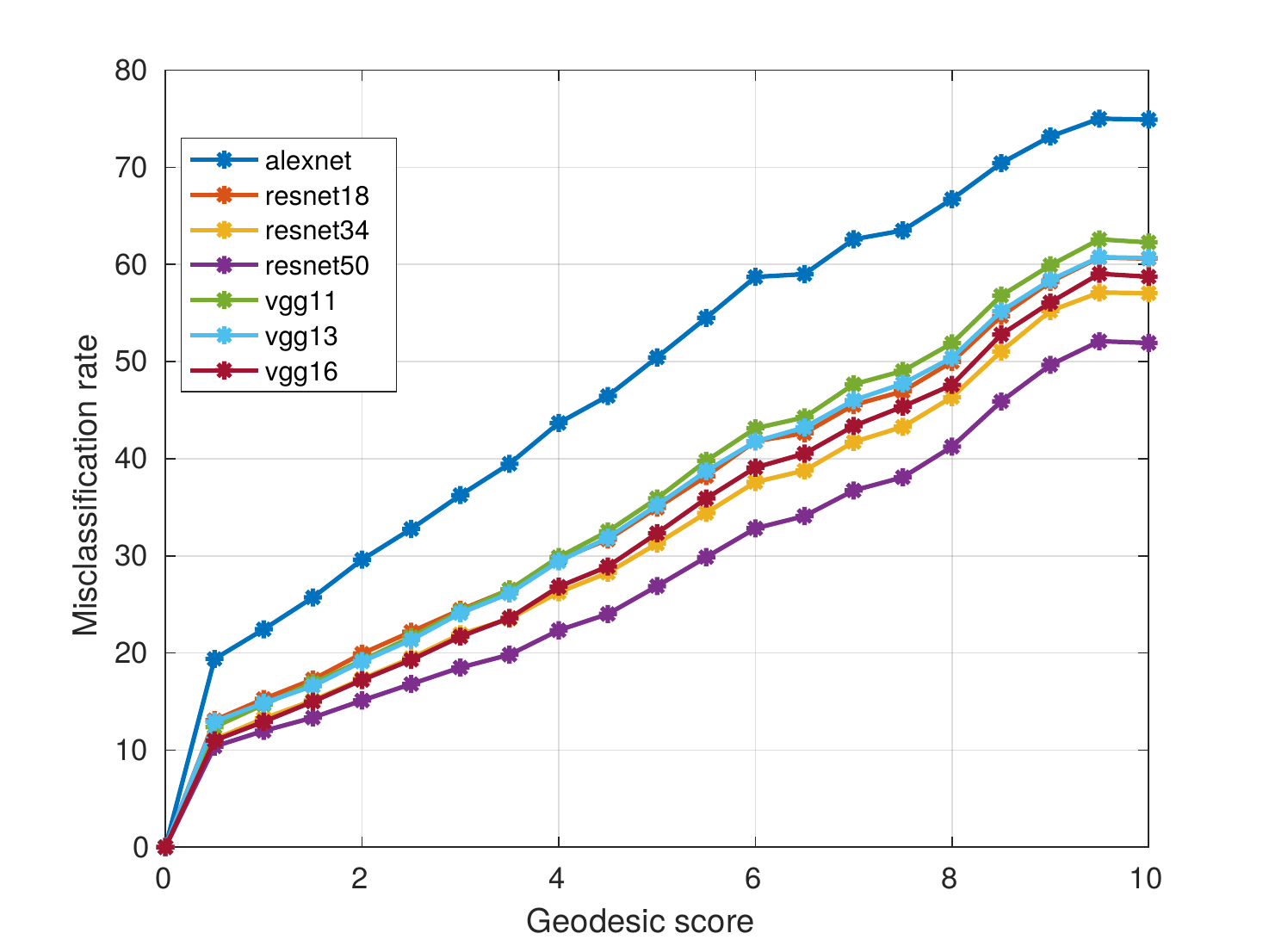}}
    		\caption{Similarity transformations}
    	\end{subfigure}
    	\hspace{5mm}
  	\begin{subfigure}[t] {0.45\textwidth}
   		\includegraphics[width=\textwidth , keepaspectratio]{{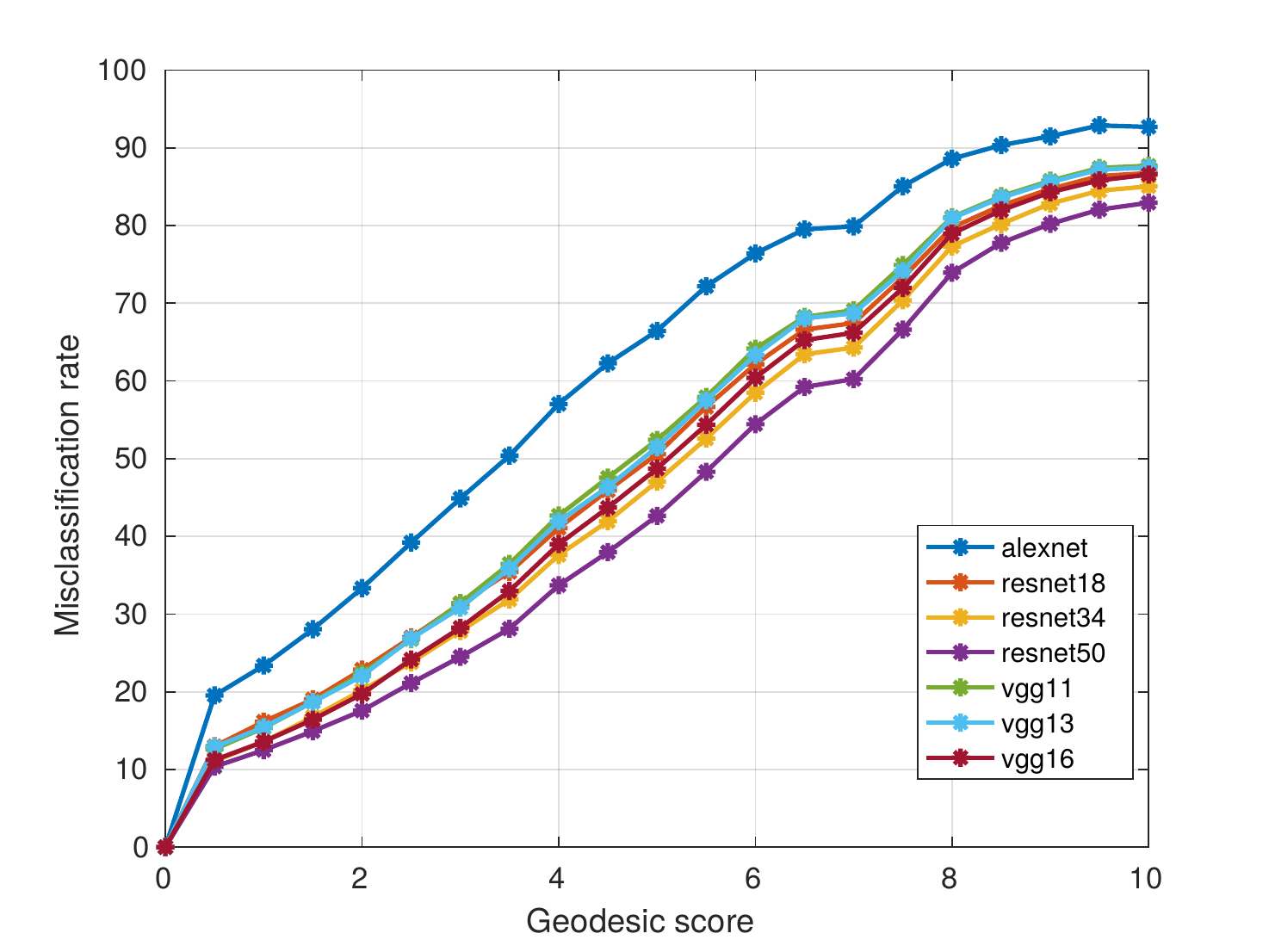}}
    		\caption{Affine transformations}
    	\end{subfigure}
    	\caption{Misclassification rates of different networks with respect to the geodesic score of the transformation of the input images. The rates are calculated using 5000 images from ILSVRC2012 validation dataset with 10 different transformations for each image.}
	\label{fig:random}
\end{figure*}
\subsection{Invariance score of different architectures}
The second experiment uses ManiFool to quantitatively compare the invariance properties of different networks. For this purpose, we have computed the invariance score of AlexNet~\cite{krizhevsky_imagenet_2012}, ResNet~\cite{resnet} and VGG~\cite{vgg} pre-trained models, using 5000 random images from ILSVRC2012 validation dataset~\cite{imagenet}.  Some examples that are created during this experiment can be seen on Figure~\ref{fig:imagenet}, which shows that these networks are quite vulnerable against geometric transformations. In addition, we also compute the invariance of these networks to random transformations for $r \in [0,10]$ using 5000 images from ILSVRC2012 validation dataset with 10 random transformations each.  The misclassification rates of the networks for each tested $r$ is seen on Figure \ref{fig:random}.
\begin{figure*}[ht!]
  \center
  \begin{subfigure}[t]{0.45\textwidth}
    \centering
    \includegraphics[width=0.8\textwidth]{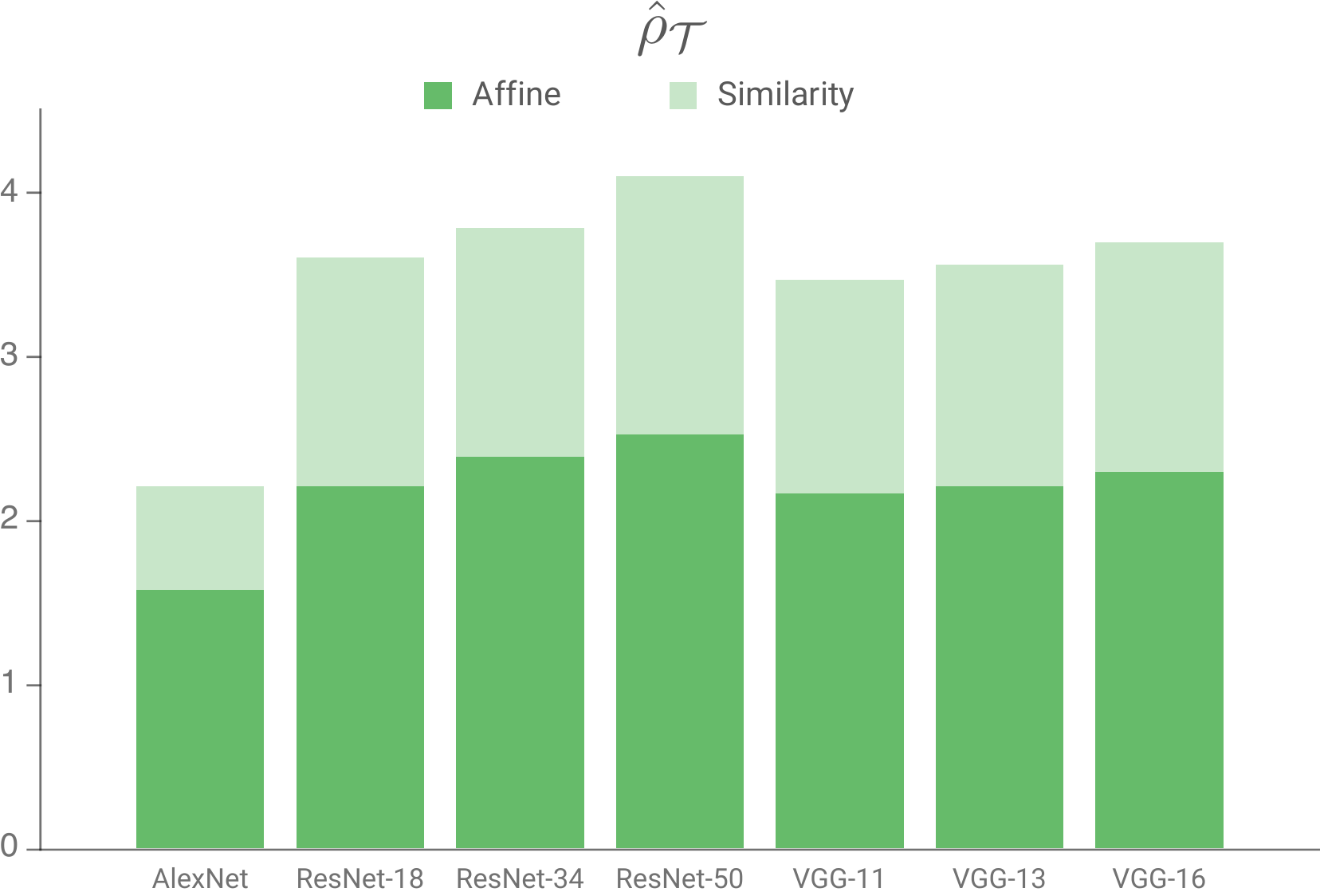}
  \end{subfigure}
  \begin{subfigure}[t]{0.45\textwidth}
    \centering
    \includegraphics[width=0.8\textwidth]{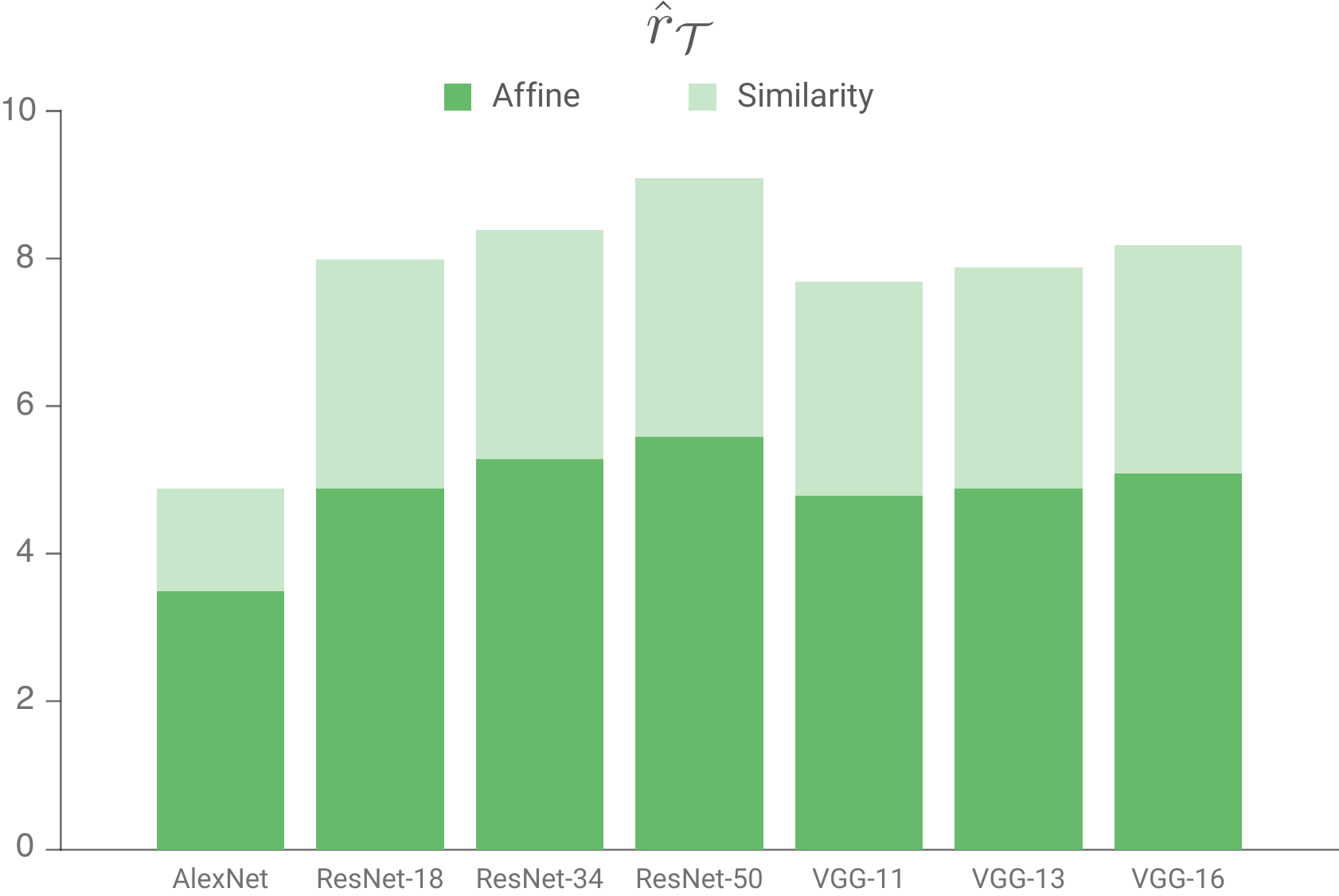}
  \end{subfigure}
  \caption{Invariance scores of different networks against similarity and affine transformations. For $\hat{\rho}_{\T}$, the invariance scores are calculated using 5000 images from ILSVRC2012 validation dataset and for $\hat{r}_{\T}$, they are calculated again using 5000 images with 10 different transformations for each image.}
  \label{fig:imnet_rho}
\end{figure*}
Figure \ref{fig:imnet_rho} reports the invariance scores for all of the networks. It can be seen that, for the same type of networks (\eg, VGGs and ResNets), the invariance increases with the number of layers in each set of transformations. This result is in agreement with the previous empirical studies on smaller datasets such as \cite{goodfellow_measuring_2009,manitest}, but we have shown here that this also holds for deeper, more complex networks. On top of this, we can also see that neither the number of parameters nor the depth of the networks are the only decisive factors: ResNet-18 is less invariant than VGG-16, even though it is deeper and VGG-16 has more parameters than ResNet-50, yet it is less invariant. Similar results can be observed for the invariance to random transformations, \eg, the invariance again increases with depth. In fact, Figure \ref{fig:imnet_rho} shows that there certainly is a correlation between these two invariance values. However, interestingly, the ordering is not exactly the same as can be seen on Figure \ref{fig:random}. For example, ResNet-34 and ResNet-50 perform better against random transformations compared to VGG networks. 
%
\subsection{Adversarial Training using ManiFool}
\begin{figure}
\centering
   		\includegraphics[width=0.4\textwidth , keepaspectratio]{{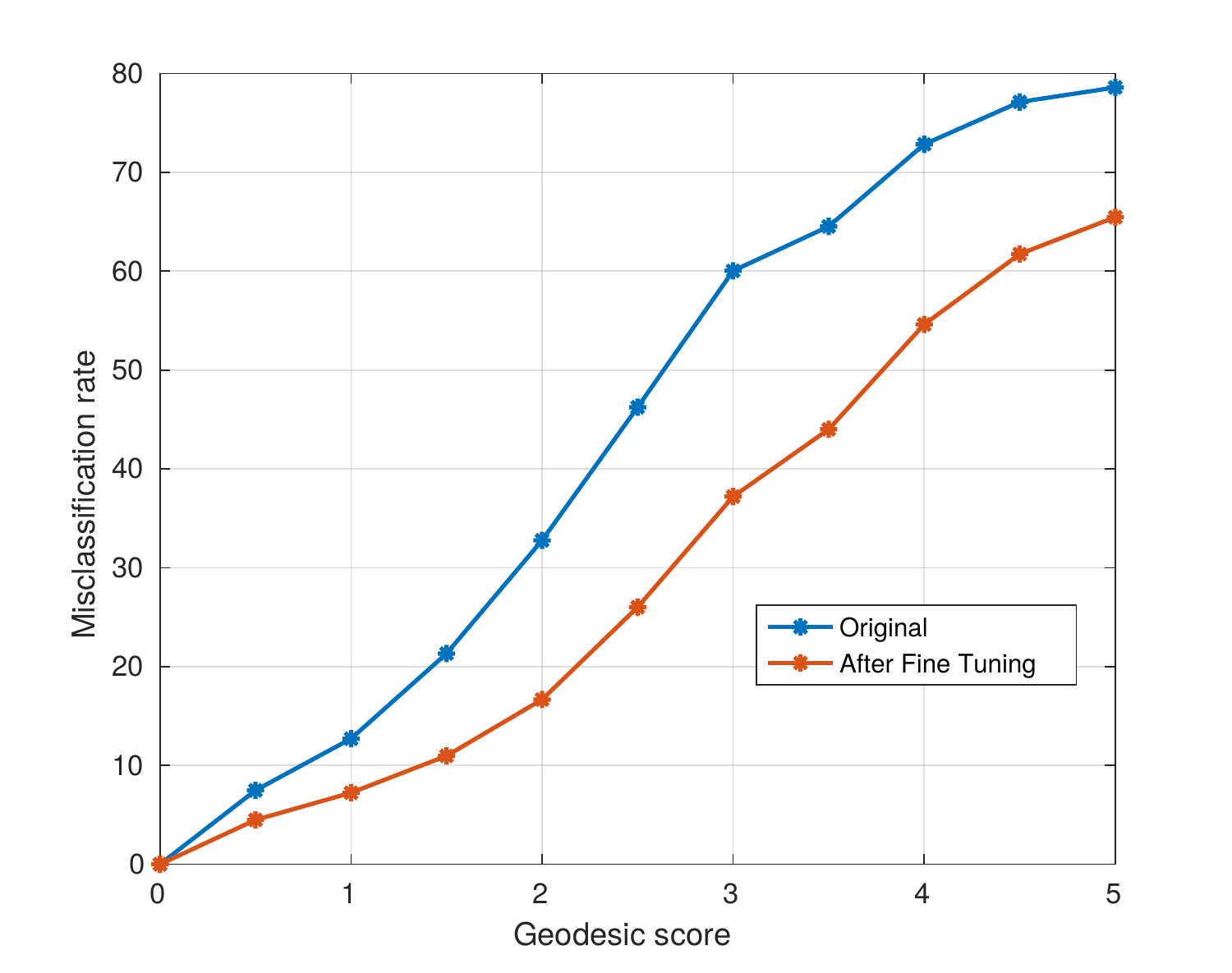}}
    		\caption{Misclassification rate of ResNet18 on CIFAR10 before and after fine tuning using adversarial geometric transformations with respect to the geodesic score of the random affine transformations of the input images. The rates are calculated using 5000 images from the test set with 20 transformations for each image.}\vspace{-5mm}
		\label{fig:adv_train_rand}
\end{figure}

\begin{table}
	\centering\small
	\begin{tabularx}{0.45\textwidth}{ |c|Y|Y|Y|Y| }
		\hline
		& Original & Minimal & Random & Baseline \\ \hline
		$\hat{\rho}_{\T}$  & 1.13 & 1.78 & 1.55 & 1.10 \\ \hline
	\end{tabularx}
	\caption{The invariance to affine transformations of ResNet18 on CIFAR10 before and after the first epoch of fine tuning. Invariance score is calculated using 5000 images from CIFAR10 test set. 'Minimal', 'Random' and 'Baseline' stand for the extra epoch done using the transformed dataset created using ManiFool, the dataset created using random transformations and the training set respectively.\vspace{-5mm}}\label{tab:ft}
\end{table}
As our last experiment, we have fine-tuned a network for CIFAR10~\cite{krizhevsky_learning_2009} classification by performing  5 additional epochs with a 50\% decreased learning rate using images that were transformed by label changing affine transformations generated using ManiFool.  To be complete, we also performed 5 extra epochs using the original data and 5 extra epochs using randomly transformed images. For these randomly transformed images, the score of the transformations were equal to the median geodesic score of the dataset generated with ManiFool. We used ResNet-18 for this experiment, which was trained using stochastic gradient descent with softmax loss. The invariance score against minimal transformations, $\hat{\rho}_{\T}$ is then calculated for the networks after each epoch using 5000 images from CIFAR-10 test set. To see the effect of fine tuning on the invariance against random transformations,  we also computed the misclassification rate for a transformation distance of $r \in [0,5]$ using 5000 images from CIFAR10 test set with 20 transformations each. The results of these experiments are seen on Figure \ref{fig:adv_train_rand} and Table \ref{tab:ft}.
It can be observed in Table \ref{tab:ft} that fine tuning with adversarial examples has increased the invariance score significantly, even after only one extra epoch. This is in line with previous works on additive adversarial perturbations, where adversarial training was shown to improve robustness against the examined additive perturbation\cite{fgs}\cite{deepfool}, but this effect is now seen for transformations as well. We can also see that its effect is greater than using only randomly transformed images. More interestingly, we can also see in Figure \ref{fig:adv_train_rand} that fine tuning using the worst-case examples has increased the robustness against \emph{random transformations} considerably. For example, for random transformations with $\tilde{d}(e,\tau_l) = 2.5$, the misclassification rate has decreased more than 20\% after fine tuning. Although there is also a small penalty in accuracy of the network (0.6\% reduction on test set), this shows  that choosing the worst-case transformation examples for fine tuning can increase invariance in both worst-case and random regimes.

\section{Conclusion}
In this work, we have presented a new constructive framework for computing the invariance score of deep image classifiers against geometric transformations. We proposed an algorithm, ManiFool, for finding small fooling transformation examples. The simple idea behind it is to perform gradient descent on the manifold of geometric transformations, in other words, it iteratively moves towards the class decision boundary while staying on the manifold to generate adversarial examples. Using this method, we have studied the robustness of networks trained on ImageNet against worst-case and random transformations. We also showed that adversarial training using ManiFool can be used as a way to improve the robustness of deep networks against both worst-case and random transformations and leads to more invariant networks. In the future, we believe this process can be used for empirical analysis of neural networks under geometric transformations and thus provide a better understanding of invariance to non-additive perturbations and the properties of different network architectures. Also, the ManiFool algorithm can be useful for generating new and practically relevant types of adversarial examples by using wider types of natural transformations.

\subsection*{Acknowledgments}
We thank Alhussein Fawzi for the fruitful discussions. We also gratefully acknowledge the support of NVIDIA Corporation with the donation of the Titan X GPU used for this research.

{\small
\bibliographystyle{ieee}
\bibliography{ThesisReferences.bib}
}

\end{document}